\documentclass[11pt]{article}
\usepackage{coling2020}
\usepackage{times}
\usepackage{url}
\usepackage{latexsym}

\usepackage{url}

\usepackage{amsmath}
\usepackage{bm}
\usepackage{color}
\usepackage{multirow}
\usepackage{subfig}
\usepackage{arydshln}
\usepackage{url}
\usepackage{amsfonts}
\usepackage{graphicx} 
\usepackage{graphics}
\usepackage{wrapfig}

\usepackage{caption}
\usepackage{balance}
\usepackage{verbatim}

\definecolor{zptu}{RGB}{18, 141, 21}

\title{Rethinking the Value of Transformer Components}

\author{
Wenxuan Wang\thanks{~~Work done when interning at Tencent AI Lab.} \\ The Chinese University of Hong Kong  
\\ {\tt jwxwang@gmail.com}  \And
Zhaopeng Tu\thanks{~~Zhaopeng Tu is the corresponding author.} \\ Tencent AI Lab \\ {\tt zptu@tencent.com}
}

\begin{document}
\maketitle
\begin{abstract}
Transformer becomes the state-of-the-art translation model, while it is not well studied how each intermediate component contributes to the model performance, which poses significant challenges for designing optimal architectures. In this work, we bridge this gap by evaluating the impact of individual component (sub-layer) in trained Transformer models from different perspectives. Experimental results across language pairs, training strategies, and model capacities show that certain components are consistently more important than the others. We also report a number of interesting findings that might help humans better analyze, understand and improve Transformer models.
Based on these observations, we further propose a new training strategy that can improves translation performance by distinguishing the unimportant components in training.
\end{abstract}

\section{Introduction}
\label{sec:intro}

\blfootnote{
    %
    %
    %
    %
    \hspace{-0.65cm}  
    This work is licensed under a Creative Commons 
    Attribution 4.0 International Licence.
    Licence details:
    \url{http://creativecommons.org/licenses/by/4.0/}.
    
    
}

Transformer~\cite{Vaswani:2017:NeurIPS} has achieved the state-of-the-art performance on a variety of translation tasks. It consists of different stacked components, including self-attention, encoder-attention, and feed-forward layers. However, so far not much is known about the internal properties and functionalities it learns to achieve the performance, which poses significant challenges for designing optimal architectures.

In this work, we bridge the gap by conducting a granular analysis of components on trained Transformer models. We attempt to understand how does each component contribute to the model outputs.
Specifically, we explore two metrics to evaluate the impact of a particular component  on the model performance: 1) {\em contribution in information flow} that manually masks individual component each time and evaluate the performance without that component; and 2) {\em criticality in representation generalization} that depends on how much closer the weights can get for each component to the initial weights while still maintaining performance. 
Those two metrics evaluate the component importance of a trained Transformer model from different perspectives.
Empirical results on two benchmarking datasets reveal the following observations (\S\ref{sec:observing}):
\begin{itemize}
    \item The decoder self-attention layers are least important, and the decoder feed-forward layers are most important.
    \item The components that are closer to the model input and output (e.g., lower layers of encoder and higher layers of decoder) are more important than components on other layers.
    \item Upper encoder-attention layers in decoder are more important than lower encoder-attention layers.
\end{itemize}
The findings are consistent across different evaluation metrics, translation datasets, initialization seeds and model capacities, demonstrating their robustness.

We further analyze the underlying reason (\S\ref{sec:analyze}), and find that lower dropout ratio and more training data lead to less unimportant components. Unimportant components can be identified at early stage of training, which are not due to deficient training. Finally, we show that unimportant components can be rewound~\cite{Frankle:2019:ICLR} to further improve the translation performance of Transformer models (\S\ref{sec:distinguish}).

\section{Methodology}
\label{sec:method}

In this section, we evaluate the importance of individual Transformer components via two different perspective: {\em contribution in information flow} and {\em criticality in representation generalization}.

\subsection{Contribution in Information Flow}

\begin{wrapfigure}{r}{0.3\textwidth}
\centering
    \includegraphics[width=0.3\textwidth]{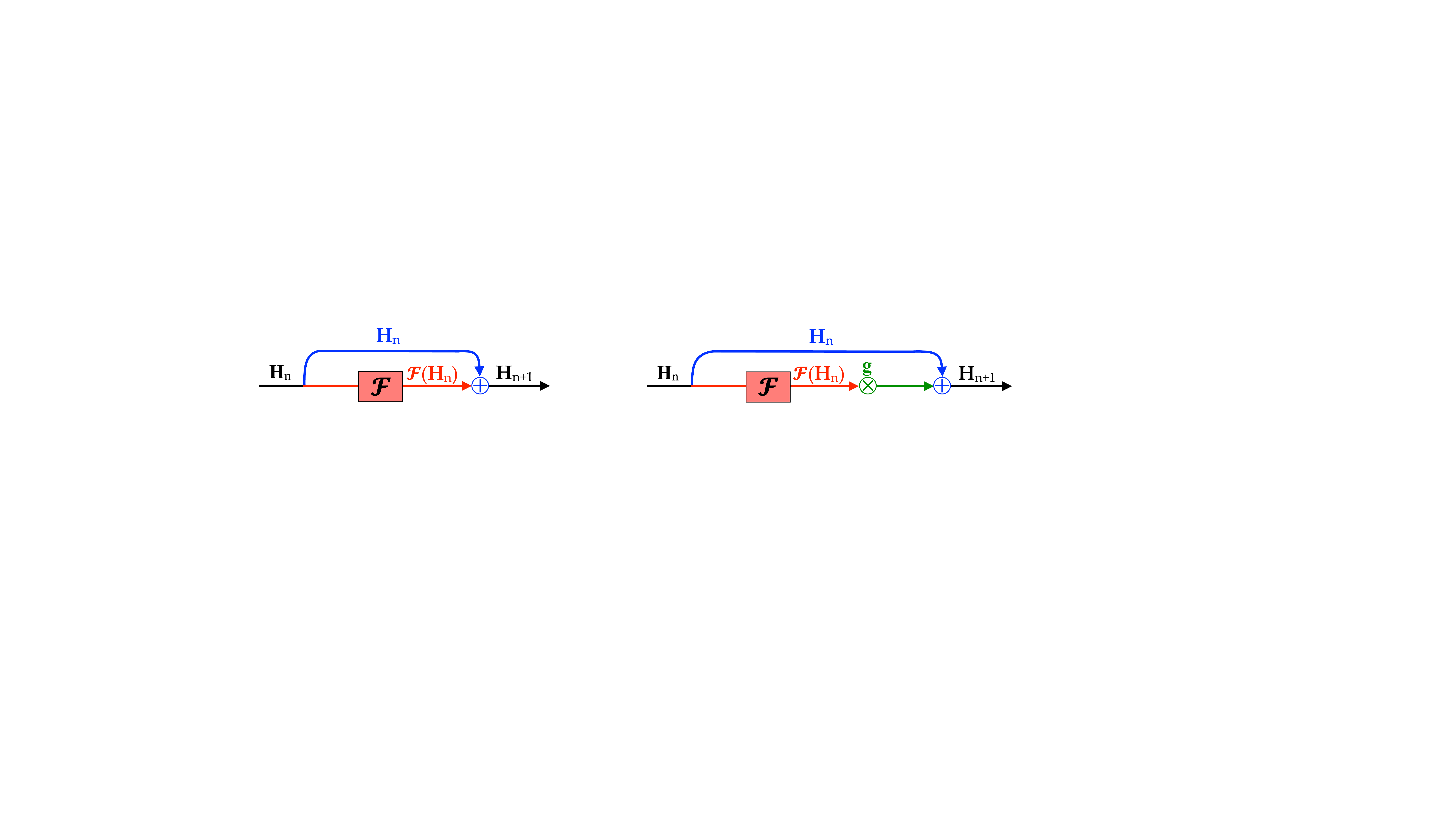} 
    \caption{Contribution of a sublayer $F$ to the information transformation.}
    \label{fig:flow}
\end{wrapfigure}
The ultimate goal of machine translation is to fully transform the information from the source side to the target side. It is essential to understand how information flows from the input, across  the encoder and the decoder, to the output. Figure~\ref{fig:flow} shows an example to illustrate how information flows across a basic Transformer component (i.e., residual sub-layer).

We first try to understand how each sub-layer contributes to the information flow from input to output.
To understand the contribution of a particular component in the information flow, we investigate the effect of masking that component.
Intuitively, we followed \newcite{Michel:2019:NeurIPS} to manually ablate each component (i.e., replacing the output with zeros) from a trained Transformer, and evaluated the performance of the resulting masked Transformer.
The component is important if the performance without that component is significantly worse than the full model's, otherwise it is redundant given the rest of the model.

Formally, we define the contribution score of $n$-th component as
\begin{equation}
    Contri_n = \frac{\widehat{M}_n}{\widetilde{M}} ~~~~~ \text{where}  ~~~~~   \widehat{M}_n = \left\{  \begin{aligned} 0 &  & {M_n < 0} \\ M_n & & {0<M_n<C} \\ C & & {M_n > C}  \end{aligned} \right., ~~ \widetilde{M} = \max\{\widehat{M}_1, \dots, \widehat{M}_N\}
\end{equation}
where $M_n$ is the BLEU drop by ablating the $n$-th component. It is first clipped to the range of $[0, C]$ to avoid the minus importance value and exploding drop. Then it is normalized to $[0,1]$ by dividing the maximum drop $\widetilde{M}$. In this study, we set the constant number $C$ as 10\% of the BLEU score of baseline model.

\subsection{Criticality in Representation Generalization}

\begin{figure}[h]
    \centering
    \subfloat[Non-critical  module]{\includegraphics[width=0.4\textwidth]{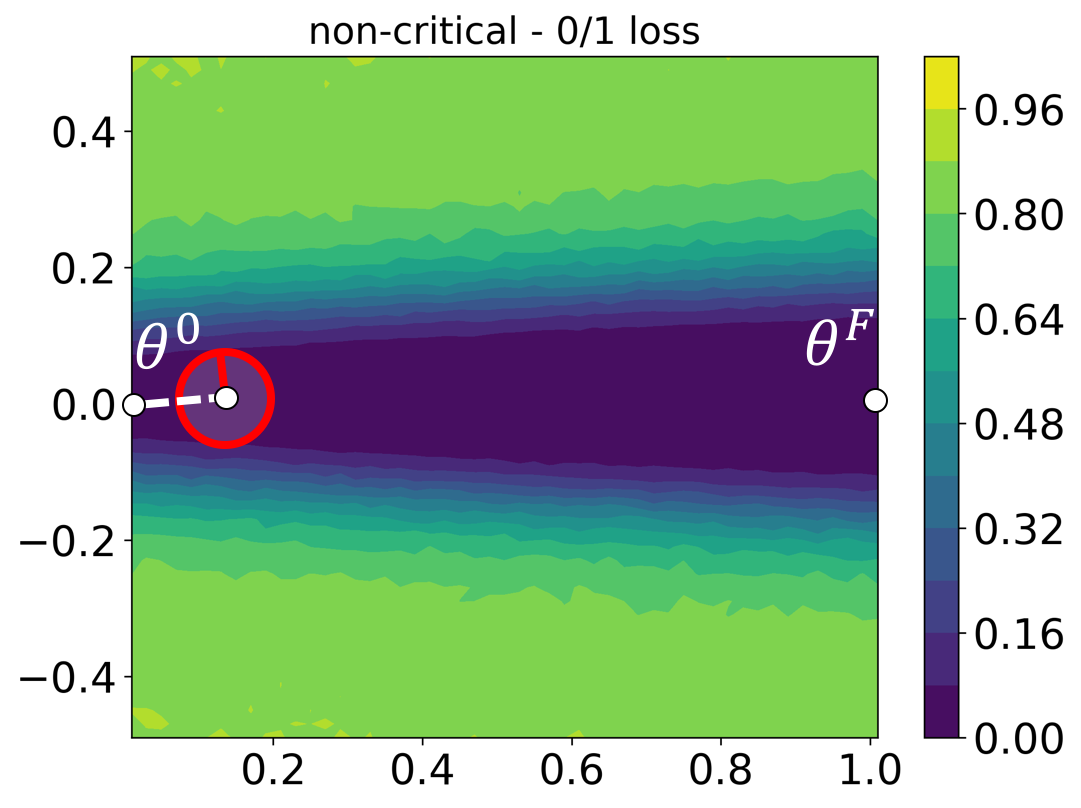}}
    \hspace{0.05\textwidth}
    \subfloat[Critical module]{\includegraphics[width=0.4\textwidth]{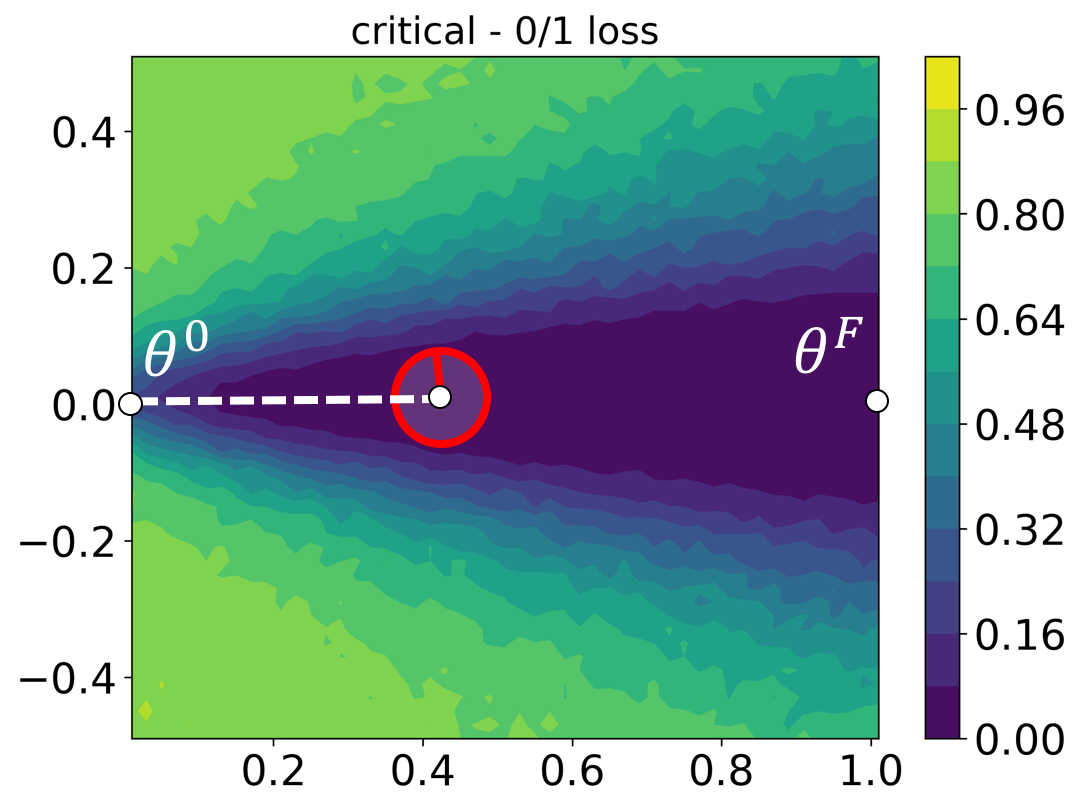}}
    \caption{Loss values in the valleys that connect the initial weights $\theta^0$ to the final weights $\theta^F$ of a non-critical (left) and a critical (right) module.
    Given a ball with radius $r$ (length of the red line), module criticality can be defined as how far one can push the ball in the valley towards initialization (length of the white dashed line) divided by the radius $r$. Figure credit:~\newcite{Chatterji:2020:ICLR}.
    }
    \label{fig:criticality}
\end{figure}

\newcite{Zhang:2019:arXiv} reported the {\em module criticality} phenomenon, in which modules of the network present different robustness characteristics to parameter perturbation (e.g. rewinding back to its initialization value). Notice that rewinding to initial value is a relaxation of setting to zeros, due to the initialization of Transformer model is Xavier Initialization with $0$ means. The module is critical if rewinding its weights to the initialization harms the network performance, otherwise it is non-critical in full network.

~\newcite{Chatterji:2020:ICLR} theoretically formulated this phenomenon and revealed that the criticality metric is reflective of the network generalization.
Specifically, they used a convex combination of the initial weights and the final weights of a module to define an optimization path to traverse.
They quantitatively defined the module criticality such that it depends on how much closer the weights can get to the initial weights on this path while still maintaining the performance. Figure~\ref{fig:criticality} shows an example. It measures how much the performance of a model rely on specific module. 

Formally, for the $n$-th component, let $\theta_n^{\alpha_n} = (1-\alpha_n)\theta_n^0 + \alpha_n \theta_n^f ,~\alpha_n \in [0,1]$ be the convex combination between initial weights $\theta_n^0$ and the final weights $\theta_n^f$. We define the criticality score of n-th component as 
\begin{equation}
 Criti_n = \min \alpha_n ~~~ s.t. ~~~ \text{BLEU}(\text{Model with }\theta_n^f) - \text{BLEU}(\text{Model with }\theta_n^{\alpha_n}) < \epsilon.
\end{equation}
In other words, criticality score is the minimum $\alpha$ to maintain the performance drop within a threshold value $\epsilon$. The criticality score of $n$-th component is small means we can move the weight of $n$-th component a long way back to initialization without hurting model performance. In this study, we use $\epsilon$ as 0.5 BLEU point, which generally indicates a significant drop of translation performance on the benchmarking datasets.

\vspace{10pt}
Although  both of the two metrics evaluate the component importance in terms of its effect on model performance, there are considerable differences. The contribution score measures the effect of fully ablating the component on model performance (i.e. {\em hard metric}), while the criticality score measures how much the component can be rewound while maintaining the model performance (i.e. {\em soft metric}).

\section{Experiments}

\paragraph{Data and Setup}

We conducted experiments on the benchmarking WMT2014 English-German (En-De) and English-French (En-Fr) translation datasets, which consist of 4.6M and 35.5M sentence pairs respectively.
We employed BPE~\cite{sennrich2016neural} with 32K merge operations for both language pairs, and used case-sensitive 4-gram NIST BLEU score~\cite{papineni2002bleu} as our evaluation metric.

Unless otherwise stated, the Transformer model consists of 6-layer encoder and decoder. The layer size is 512, the size of feed-forward sub-layer is 2048, and the number of attention heads is 8. We followed the settings in~\cite{Vaswani:2017:NeurIPS} to train the Transformer models on the En-De and En-Fr datasets. We set the dropout as $0.1$ and the initialization seed as $1$ for all Transformer models .

\subsection{Observing Component Importance}
\label{sec:observing}

In this section, we first measure the component importance of trained Transformer models. Then we vary some settings, which are threats to validity, to verify the consistency of our finding.

\begin{figure}[h]
\centering
    \subfloat[{En-De}: Contribution]{
    \includegraphics[width=0.2\textwidth]{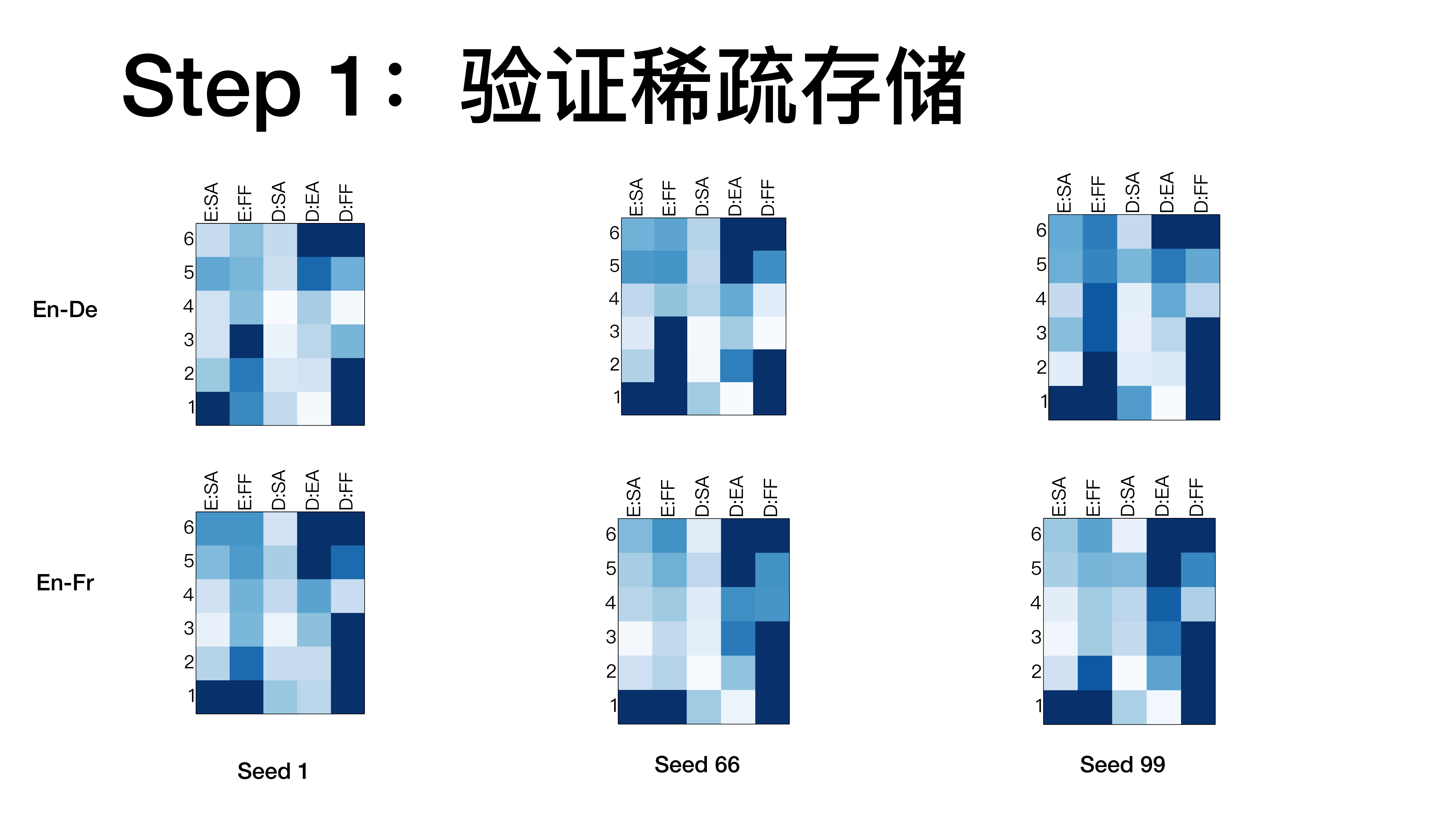}} \hspace{0.01\textwidth}
    \subfloat[{En-Fr}: Contribution]{
    \includegraphics[width=0.2\textwidth]{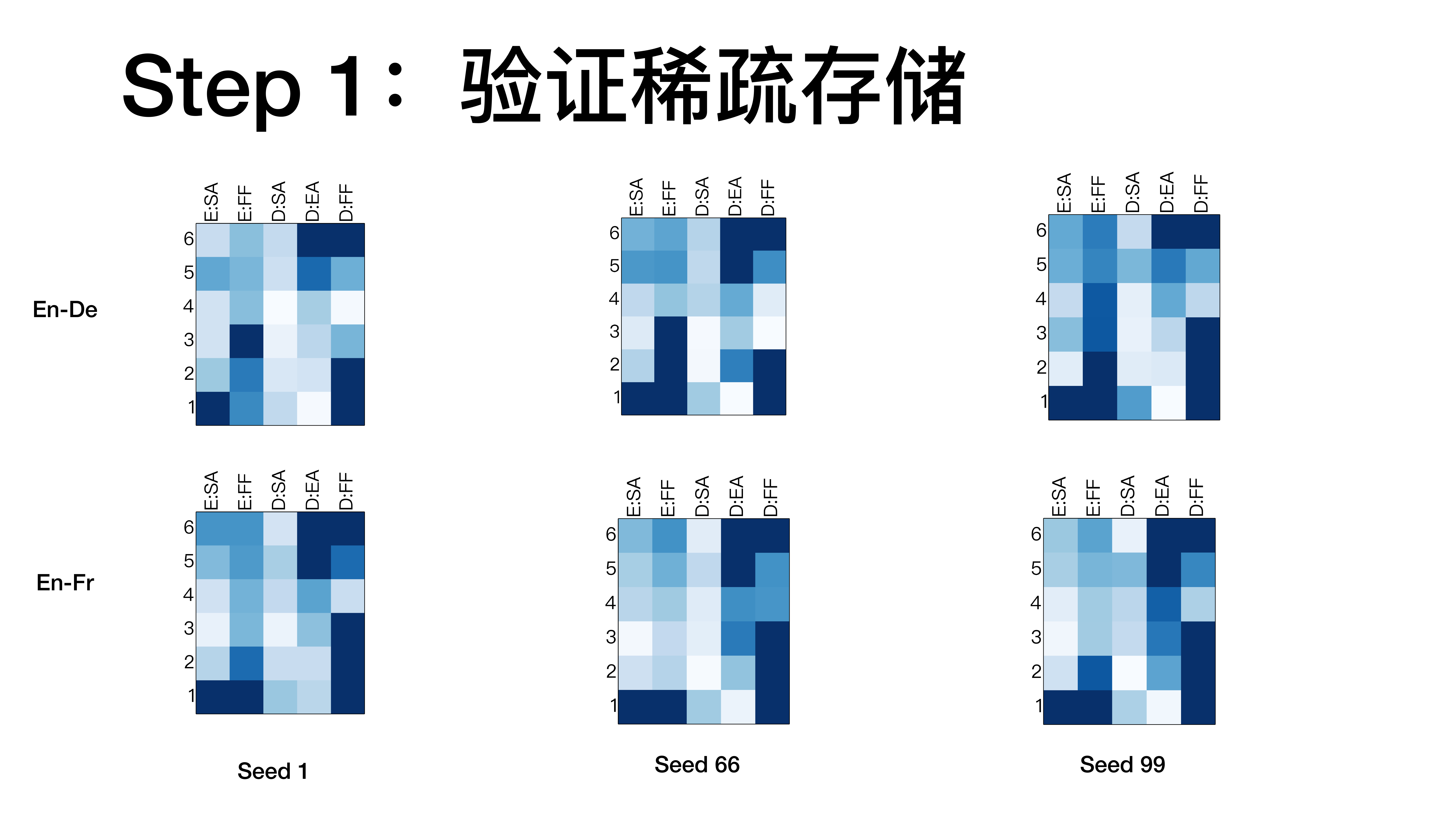}}
    \hspace{0.05\textwidth}
    \subfloat[{En-De}: Criticality]{
    \includegraphics[width=0.2\textwidth]{figures/criti-en-de.pdf}}
    \hspace{0.01\textwidth}
    \subfloat[{En-Fr}: Criticality]{
    \includegraphics[width=0.2\textwidth]{figures/criti-en-fr.pdf}}
    \caption{Importance of individual components measured by (a, b) contribution in information flow, (c, d) criticality in representation generalization. Y-axis is the layer id and X-axis is the type of components. "E", "D", "SA", "EA" and "FF" represent Encoder, Decoder, Self-attention, Encoder-attention and Feed-forward layer respectively. {\em Darker cells denote more important components.}}
    \label{fig:layer-importance}
\end{figure}

\paragraph{Several observations on component importance.} Figure~\ref{fig:layer-importance} shows the importance of Transformer components measured by two metrics. The two importance metrics agree well with each other, and reveal several observations in common:
\begin{itemize}
    \item In general, the decoder self-attention layers (``D:SA'') are least important, and the decoder feed-forward layers (``D:FF'') are most important.
    \item Lower components in encoder (e.g. ``E:SA'' and ``E:FF'') and higher components  in decoder (e.g. ``D:EA'' and ``D:FF'') are more important. This is intuitive, since these components are closer to the input and output sequences, thus are more important for input understanding and output generation.
    \item Higher encoder-attention (``D:EA'') layers in decoder are more important than lower encoder-attention layers. This is the same in \newcite{Voita:2019:ACL} which claims that lower part of decoder is more  like a language model. For the other components, the bottom and top layers are more important than the intermediate layer.
\end{itemize}

 We notice the main difference between the results of two metrics is on bottom feed-forward layers in decoder. The contribution score is high but criticality score is low. It is because the performance are bad when $\alpha = 0$ and $1$, but the performance are dramatically good when $\alpha \geq 3$. So the contribution is high but criticality is relatively low, according to the definition in Section \ref{sec:method}.

In the following experiments, we discuss the threats to validity that could affect our finding. Unless otherwise stated, we use contribution score as the default importance metric and report results on the En-De dataset.


\begin{figure}[h]
\centering
    \subfloat[Seed 66]{
    \includegraphics[width=0.2\textwidth]{figures/en-de-seed66.pdf}} \hspace{0.01\textwidth}
    \subfloat[Seed 99]{
    \includegraphics[width=0.2\textwidth]{figures/en-de-seed99.pdf}}
    \hspace{0.05\textwidth}
    \subfloat[Deep Tran.]{
    \includegraphics[width=0.12\textwidth]{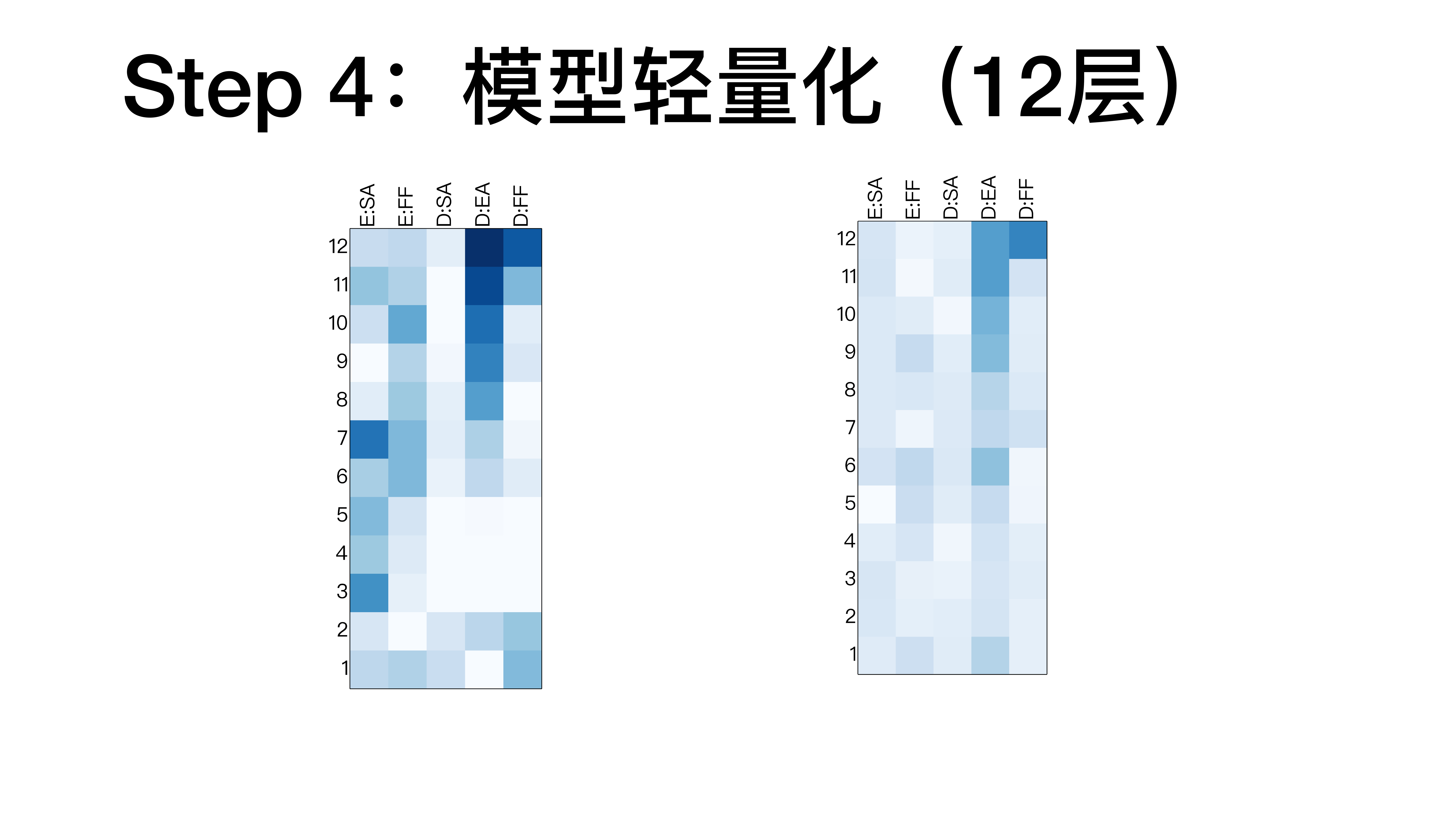}}
    \hspace{0.01\textwidth}
    \subfloat[Big Transformer]{
    \includegraphics[width=0.2\textwidth]{figures/big-en-de.pdf}} 
    \caption{Component importance for Transformer models with (a, b) different initialization seeds, and (c, d) different model capacities on the En-De dataset.}
    \label{fig:factors}
\end{figure}

\paragraph{Consistency across different initialization seeds and model capacities.}
As aforementioned, we evaluate the component importance based on a trained NMT model, which can be influenced by various hyper-parameters. We identify two hyper-parameters that have been reported to significantly influence the model performance:
\begin{itemize}
    \item {\em Initialization Seed}: Recent works have shown that neural models are very sensitive to the initialization seeds: even with the same hyper-parameter values, distinct random seeds can lead to substantially different results~\cite{Dodge:2020:arXiv}.
    \item {\em Model Capacity}: Depth and width are two key aspects in the design of a neural network architecture. ~\newcite{Lu:2017:NIPS} claimed that the depth of a network may determine the abstraction level, and the width may influence the loss of information in the forwarding pass.
    Recent studies have also demonstrated the significant effect of varying depth~\cite{Wang:2019:ACL} and width~\cite{Vaswani:2017:NeurIPS} on NMT models.
\end{itemize}
Figure~\ref{fig:factors} shows the results of Transformer models with different initialization seeds and model capacities on the En-De dataset. Specifically, we used two other different initialization seeds (i.e., ``66'' and ``99''). For the model capacity setting, we used  deeper Transformer (i.e., 12 layer) and wider Transformer (i.e., layer size be 1024). Clearly, the above conclusions hold in all cases, demonstrating the robustness of our findings.
In the following experiments, we use Transformer-base with initialization seed 1 as the default model.

\begin{wrapfigure}{r}{0.28\textwidth}
\centering
    \vspace{-25pt}
    \includegraphics[width=0.2\textwidth]{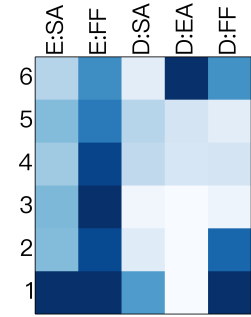} 
    \caption{Component Importance for Transformer with LayerDrop.}
    \vspace{-15pt}
    \label{fig:layerout}
\end{wrapfigure}
\paragraph{Results on Transformer trained with structured dropout.}
The Transformer model is trained without being aware of subsequent layer-wised ablating, which potentially affects the validity of our conclusions. In response to this problem, we followed~\newcite{Fan:2020:LCLR} to explore {\em LayerDrop}, a form of structured dropout, which has a regularization effect during training and allows for efficient pruning at inference time.
LayerDrop randomly drops entire components during training, which has the advantage of making the network robust to subsequent pruning.
Figure~\ref{fig:layerout} depicts the component importance of Transformer trained with LayerDrop, which reconfirms our claim that different components at different layers make distinct contributions to the model performance.

\subsection{Analyzing Unimportant Components}
\label{sec:analyze}

In this section, we delve into further analysis of unimportant components. We first find several factors that can affect the number of unimportant components. Then we attempt to find the reason of the existence  of unimportant components by representation similarity analysis, learning dynamic analysis and Layerwise Isometry Check.

\begin{figure}[h]
\centering
    \subfloat[Dropout: 0.0]{
    \includegraphics[width=0.2\textwidth]{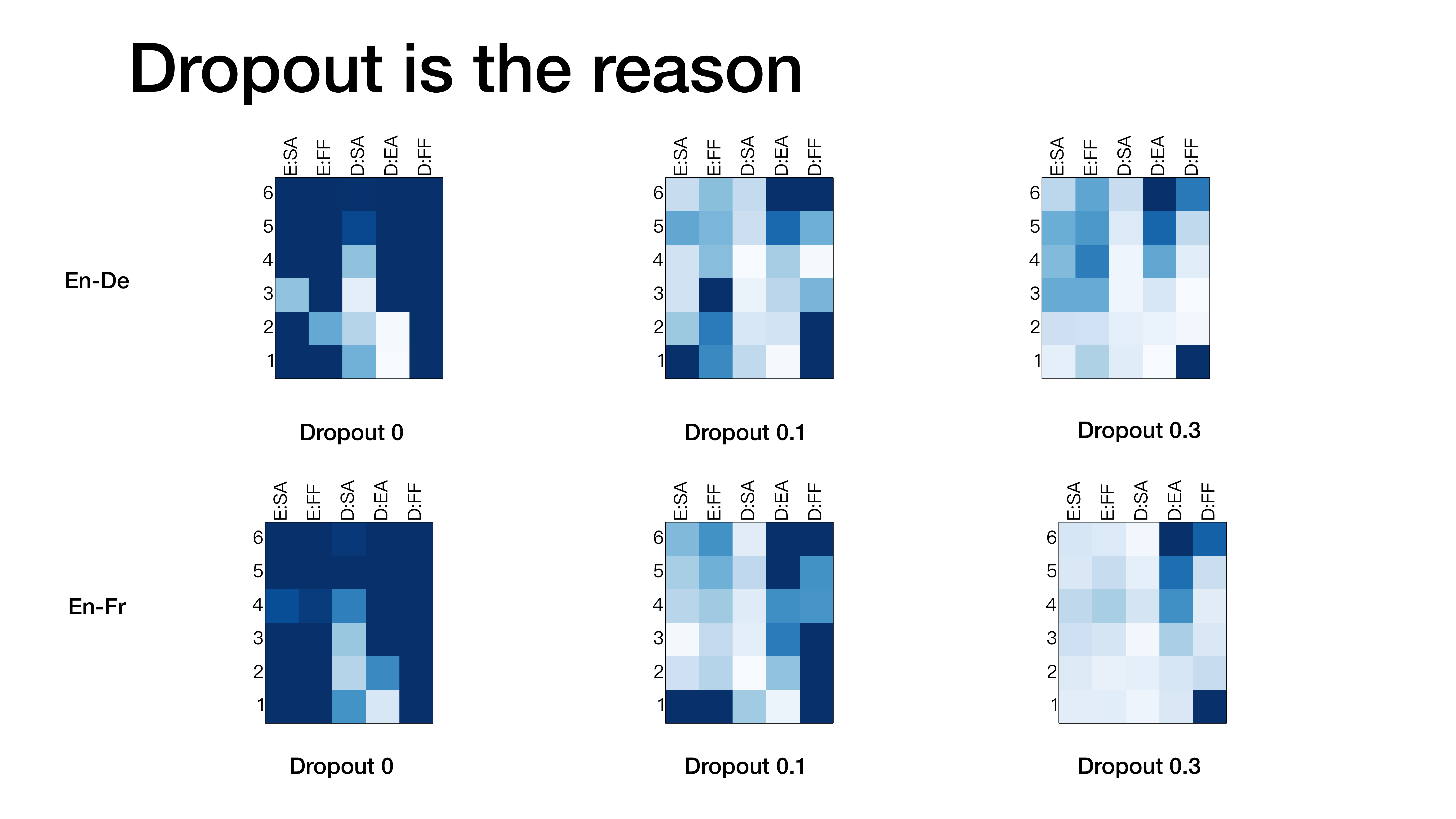}} \hspace{0.02\textwidth}
    \subfloat[Dropout: 0.1]{
    \includegraphics[width=0.2\textwidth]{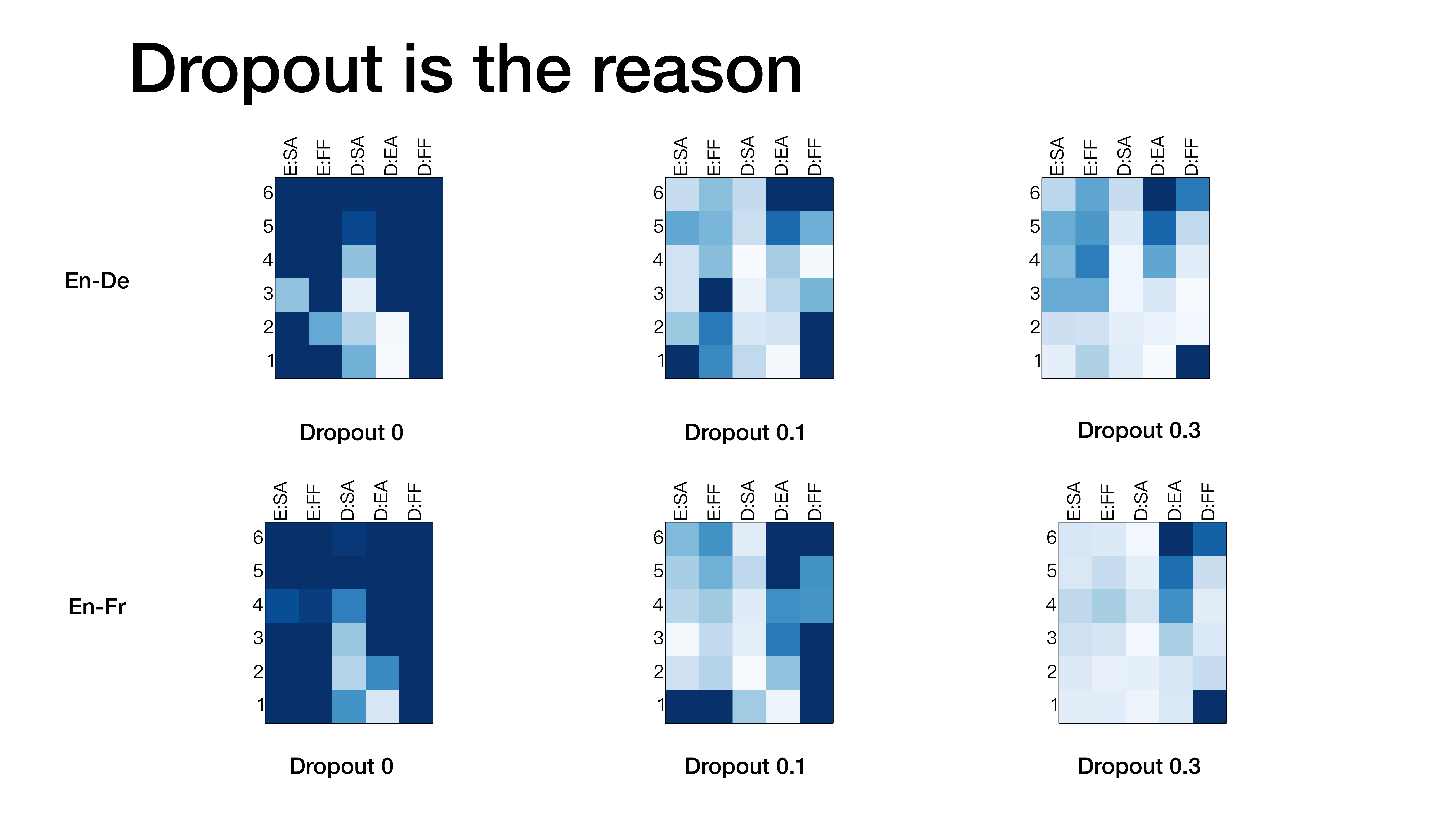}}
    \hspace{0.02\textwidth}
    \subfloat[Dropout: 0.3]{
    \includegraphics[width=0.2\textwidth]{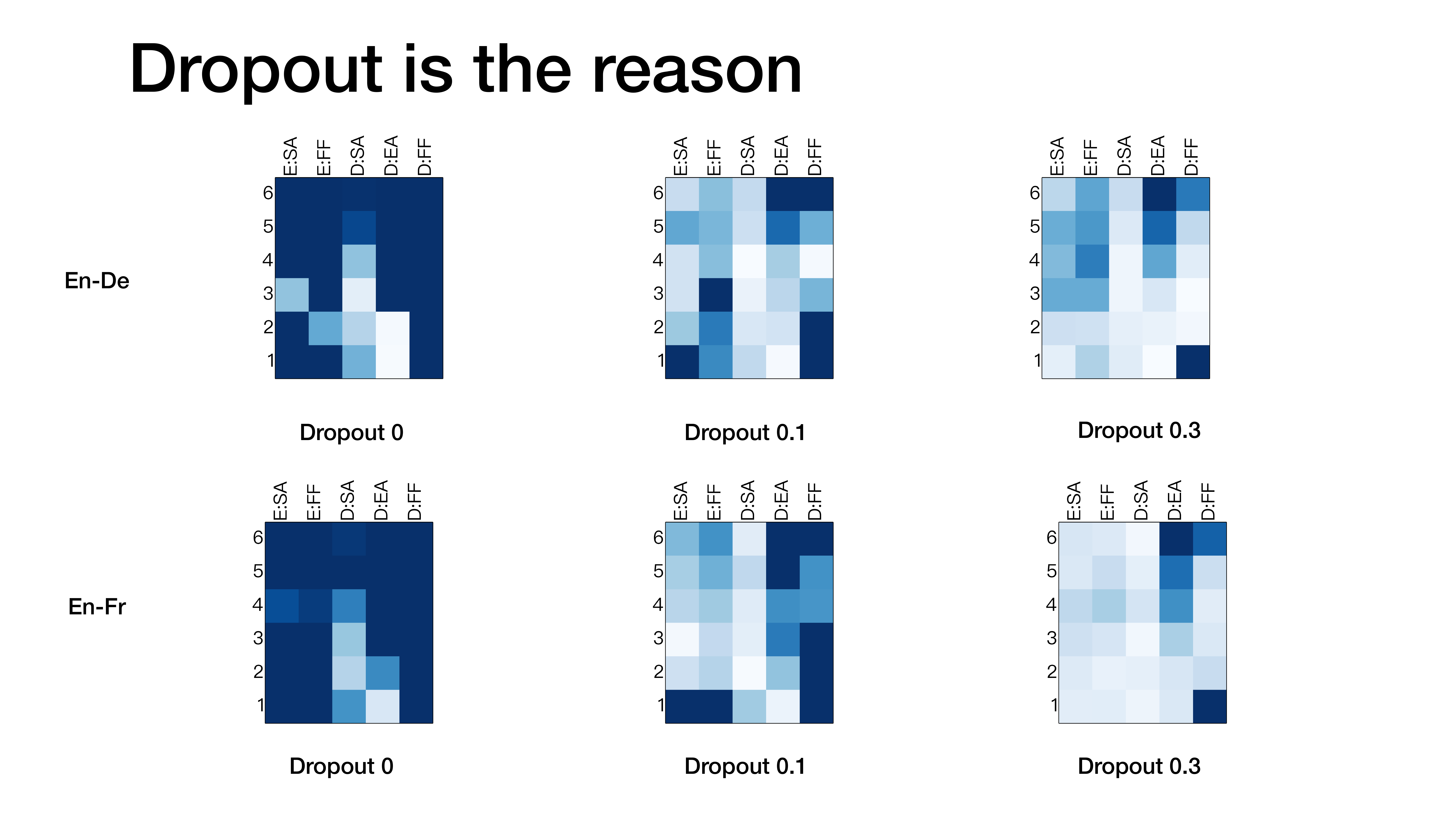}} \hspace{0.02\textwidth}
    \subfloat[Dropout: 0.5]{
    \includegraphics[width=0.2\textwidth]{figures/dropout-05-en-de.pdf}}
    \caption{Effect of dropout ratio on component importance on the En-De dataset.}
    \label{fig:en-de-dropout}
\end{figure}

\paragraph{Lower dropout ratio and more training data lead to less unimportant components.}
The training procedures of neural networks have rapidly evolved in recent years. In the experiments, we identify some factors that wound affect  the number of important components:
\begin{itemize}
    \item {\em Dropout}~\cite{Hinton:2012:arXiv}: Dropout is a commonly used technology to avoid over-fitting by randomly dropping model weights with a specific probability. In order to maintain the functionality, the model trained with dropout tends to have certain redundancy, which may explain our observation that some components can be pruned without the degrade of performance (i.e., unimportant components).
    \item {\em Training Data Size}: Larger-scale training data generally contains more patterns, which may require more components of the Transformer model to learn (i.e., important components).
\end{itemize}

Figure~\ref{fig:en-de-dropout} shows the effect of dropout ratio on component importance. We varied the dropout ratio in [0.0, 0.1, 0.3, 0.5] and trained different Transformer models with different dropout ratios from scratch on the En-De dataset. The BLEU scores are $25.58$, $27.56$, $27.43$ and $25.72$ respectively. Generally, the lower the dropout ratio, the fewer number of unimportant components the model has. One possible reason is that higher dropout ratio generally makes the trained model have more redundant components to accomplish the same functionality, thus more components can be pruned without degrading the model performance (i.e., unimportant components).

\begin{figure}[h]
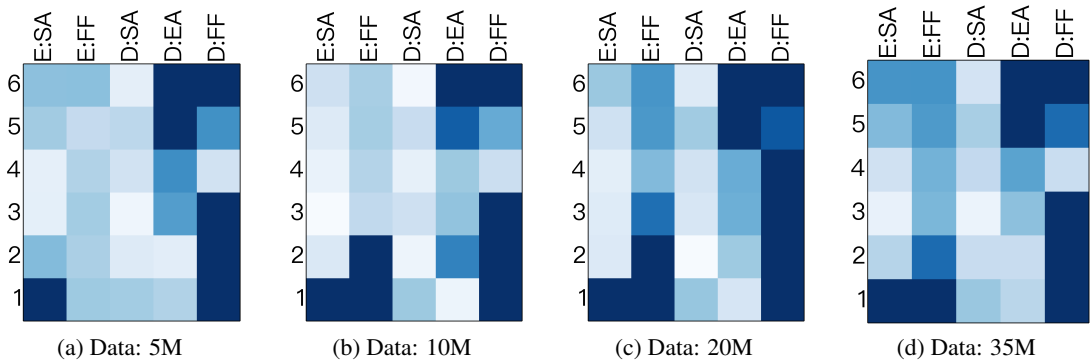

\centering
    \subfloat[Data: 5M]{
    \includegraphics[width=0.2\textwidth]{figures/en-fr-5m-ablate.pdf}} \hspace{0.02\textwidth}
    \subfloat[Data: 10M]{
    \includegraphics[width=0.2\textwidth]{figures/en-fr-10m-ablate.pdf}}
    \hspace{0.02\textwidth}
    \subfloat[Data: 20M]{
    \includegraphics[width=0.2\textwidth]{figures/en-fr-20m-ablate.pdf}} \hspace{0.02\textwidth}
    \subfloat[Data: 35M]{
    \includegraphics[width=0.2\textwidth]{figures/contri-en-fr.pdf}}
    \caption{Effect of training data size on component importance on the En-Fr dataset.}
    \label{fig:en-fr-data-size}
\end{figure}

Figure~\ref{fig:en-fr-data-size} shows the effect of different training data sizes on component importance. We randomly sampled $5M$, $10M$, $15M$, $20M$ examples from the En-Fr dataset, and trained different Transformer models on different subset. The BLEU scores are $39.67$, $39.94$, $40.44$ and $40.71$ respectively. As seen, the more training data, the more important components are required, which confirms our hypothesis.

In all cases, the lowermost E:SA and D:FF components, as well as the uppermost D:EA component are identified as important, which is consistent with the findings in Section~\ref{sec:observing}.

\begin{wraptable}{r}{0.41\textwidth}
  \centering
  \vspace{-10pt}
  \begin{tabular}{c|c|c}
  \bf Component    &   \bf En-De   &   \bf En-Fr\\
  \hline
  Important          &   0.54   &   0.55\\
  \hline
  Unimportant        &   0.42   &   0.43\\
  \end{tabular}
  \caption{Representation similarity between components output and model output.}
  \vspace{-10pt}
  \label{tab:similarity}
\end{wraptable}

\paragraph{Unimportant components outputs are less similar to the output layer representation.} In order to understand why ablating unimportant layers doesn't harm the translation performance, we used Projection Weighted Canonical Correlation Analysis (PWCCA)~\cite{Morcos:2018:NeurIPS} to analyze the similarity of layer-wise representation between each component output and final output. Table \ref{tab:similarity} shows the similarity results. We averaged the similarity score of Top-7 most important layers (listed in Figure \ref{fig:ablation}(b,c)) and Top-7 most unimportant layers. The representations of unimportant components  are less similar to the output layer representation, comparing with important components'.

\begin{figure}[h]
\centering
\includegraphics[width=0.95\textwidth]{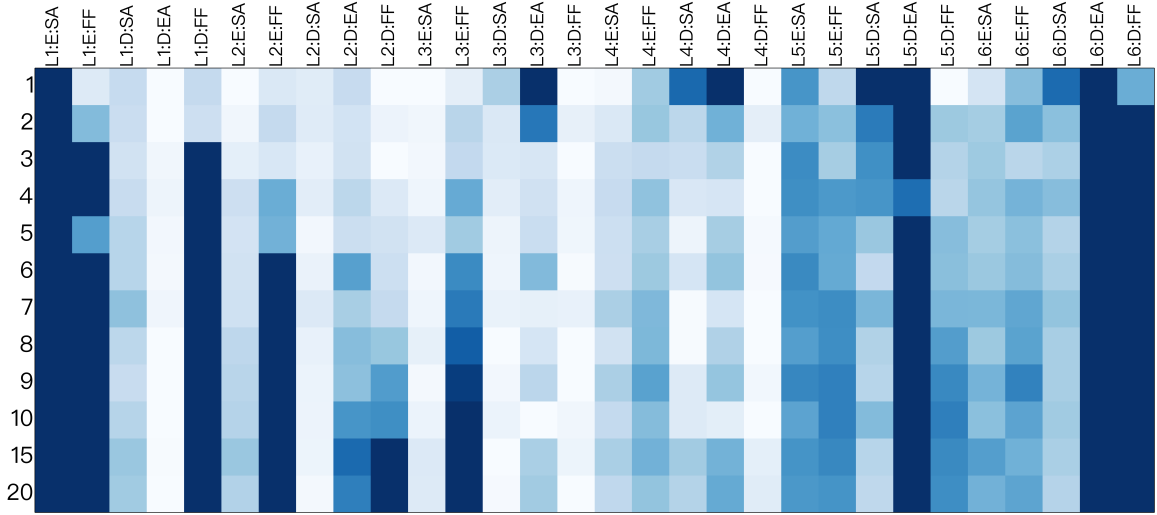}     
\caption{Learning dynamics of component importance on the En-De dataset.}
\label{fig:learning}
\end{figure}

\paragraph{Unimportant components can be identified at early stage of training.}
Recent studies have revealed that unimportant weights in a dense model can be identified at early stage of training~\cite{You2020DrawingET}. 
\newcite{Lee2020ASP} further claimed that the initialization value decides the unimportant weights.
Inspired by these findings, we try to answer the question: {\em are unimportant components created to be unimportant?}
Figure~\ref{fig:learning} illustrates the learning dynamics of component importance on the En-De dataset. Although most of the important components can be identified at early stage of training (e.g., epoch 3 or 4), they cannot be identified at initialization. The finding is also consistent with the similar important components of different initialization seeds (Figures~\ref{fig:factors} (a, b)).

\begin{table}[h]
  \centering
  \begin{tabular}{c||cc|ccc||cc|ccc}
  \multirow{2}{*}{\bf Layer}   &   \multicolumn{5}{c||}{\bf En-De}  &   \multicolumn{5}{c}{\bf En-Fr}\\
  \cline{2-11}
    &   \em E:SA    &   \em E:FF    &   \em D:SA    &   \em D:EA  &   \em D:FF   &   \em E:SA    &   \em E:FF    &   \em D:SA    &   \em D:EA  &   \em D:FF \\
  \hline
  \hline
  6 &   1.44    &   1.32    &   0.15    &   0.14    &   1.34   &   1.44    &   1.36    &   0.14  &    0.15    &   1.36\\
  5 &   1.44    &   1.34    &   0.17    &   0.14    &   1.35   &   1.44    &   1.37    &   0.18  &    0.15    &   1.36\\
  4 &   1.46    &   1.34    &   0.22    &   0.14    &   1.33   &   1.46    &   1.36    &   0.25  &    0.15    &   1.35\\
  3 &   1.45    &   1.34    &   0.29    &   0.14    &   1.32   &   1.46    &   1.36    &   0.25  &    0.15    &   1.35\\
  2 &   1.45    &   1.34    &   0.40    &   0.15    &   1.32   &   1.46    &   1.36    &   0.37  &    0.15    &   1.35\\
  1 &   1.71    &   1.34    &   0.93    &   0.15    &   1.32   &   1.69    &   1.36    &   0.98  &    0.15    &   1.35\\
  \end{tabular}
  \caption{Results of layerwise dynamical isometry check.}
  \label{tab:ldi}
\end{table}

\paragraph{Unimportant components are not due to deficient training.} Some researchers may doubt that a component fails to contribute to the model performance (i.e., ``unimportant'') since it is not fully trained. \newcite{Lee2020ASP} claimed when the initial weights are not chosen appropriately, the propagation of input signals into layers can result in saturating error signals (i.e., gradients) under back propagation, leading to training deficiency. To dispel the doubt, we check whether a component can be efficiently trained by analyzing the signal propagation properties.

\newcite{Saxe:2014:ICLR} introduced dynamical isometry to measure a faithful signal propagation, in which signals propagate in a network isometrically with minimal amplification or attenuation.~\newcite{Lee2020ASP} showed that a sufficient condition to ensure faithful propagation is {\em layerwise dynamical isometry}, which is defined as singular values of the layerwise Jacobians being concentrated around 1.
This can guarantee that the signal from layer $n$ is propagated to layer $n-1$ (or vice versa) without amplification or attenuation in any of its dimension, which leads to efficient update of parameters of the corresponding component.

Table~\ref{tab:ldi} lists the results of layerwise dynamical isometry check. Each type of components in different layers have similar layerwise isometry values, which cannot explain their different importance on the model performance. This indicates that the existence of unimportant components are not due to deficient training (i.e. unfaithful signal propagation). The results on decoder self-attention components are different because of the existence of masking.

\subsection{Distinguishing and Utilizing Group of Unimportant components}
\label{sec:distinguish}

In our previous experiments, we observed the effect of ablating one sub-layer each time, without considering what would happen if we ablate more layers at the same time. In this section, we first identify a group of unimportant components from a trained Transformer model, and then investigate how to exploit them to improve translation performance.

\begin{figure}[h]
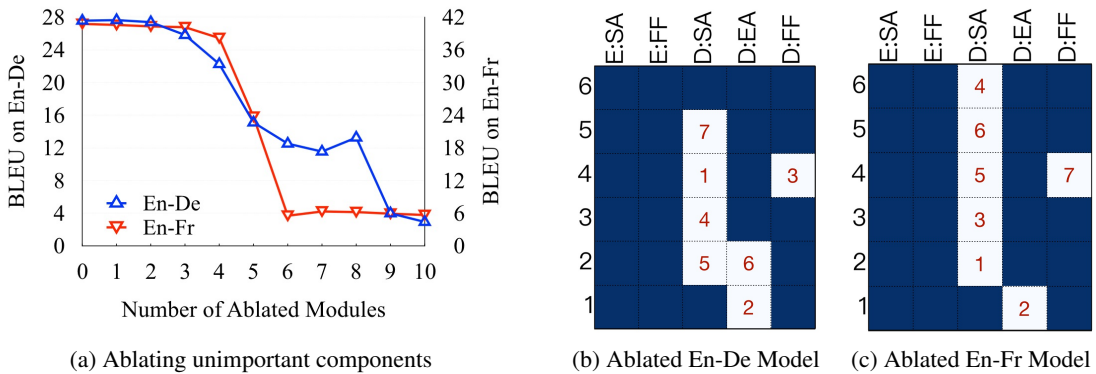

    \centering
    \subfloat[Ablating unimportant components]{\includegraphics[width=0.45\textwidth]{figures/ablation-curves.pdf}}
    \hspace{0.03\textwidth}
    \subfloat[Ablated En-De Model]{\includegraphics[width=0.21\textwidth]{figures/en-de-mask-order.pdf}}
    \hspace{0.01\textwidth}
    \subfloat[Ablated En-Fr Model]{\includegraphics[width=0.21\textwidth]{figures/en-fr-mask-order.pdf}}
    \caption{(a) Translation performance of iteratively ablating unimportant components from trained Transformer models. (b, c) The resulting ablated models on the En-De and En-Fr datasets. Number denote ablating order.}
    \label{fig:ablation}
\end{figure}

\paragraph{Identify a group of unimportant components from a trained model.} We first followed~\newcite{Michel:2019:NeurIPS} to iteratively ablate multiple components from a trained Transformer model, and report the BLEU score of ablated model (without retraining) in Figure~\ref{fig:ablation}(a). Although a few unimportant components (e.g., 3 or 4 components) can be ablated together without performance drop, ablating more components significantly harms the translation performance. These results reconfirm our analysis on the redundancy of components in Section \ref{sec:analyze}. For example, suppose two components $A$ and $B$ are considered redundant, individually ablating one of them does not harm model performance. However, it does not necessarily mean ablating both of them would not harm the performance as well. 

Figures~\ref{fig:ablation}(b, c) list the identified group of unimportant components in Transformer models trained on the En-De and En-Fr datasets. Specifically, we ablated 7 most unimportant components (i.e., 20\% components) that can harm the model performance most.  In the following experiments, we utilize unimportant components to improve translation performance with two strategies, namely {\em components pruning} and {\em components rewinding}.

\begin{wraptable}{r}{0.5\textwidth}
  \centering
  \vspace{-10pt}
  \begin{tabular}{c||r|c||r|c}
  \multirow{2}{*}{\bf Model}    &   \multicolumn{2}{c||}{\bf En-De}    &   \multicolumn{2}{c}{\bf En-Fr}\\
  \cline{2-5}
        &   \em Para.    &  \em BLEU    &   \em Para.    &   \em BLEU \\
  \hline
  \hline
  Standard          &   98M     &   27.56   &   95M  &   40.75\\
  Shallow           &   89M     &   27.31   &   86M  &   40.12\\
  \hline
  Pruned            &   89M     &   27.60   &   86M  &   40.40\\ 
  \end{tabular}
  \caption{Translation performance of pruning unimportant components.  ``Shallow'' denotes a 4-layer decoder model, which has similar number of parameters with the pruned model. All models are trained from scratch with the same hyper-parameters.}
  \vspace{-10pt}
  \label{tab:prune_bleu}
\end{wraptable}

\paragraph{Prune unimportant components and retrain the model.}
Since some of the layers are consistently unimportant, we built a model without those unimportant components and trained it from scratch (denoted as pruned model). Table~\ref{tab:prune_bleu} lists the translation performance of the pruned model. Since the pruned unimportant components are all distributed in the decoder side, we also implemented Transformer model with shallower decoder, which has the same number of parameters with the pruned model. As seen, the pruned model achieves competitive performance with the standard Transformer, and consistently outperforms the shallow model, demonstrating the reasonableness of the identified unimportant components. 


\begin{wraptable}{r}{0.5\textwidth}
  \centering
  \vspace{-10pt}
  \begin{tabular}{l||r|c||r|c}
  \multirow{2}{*}{\bf Training}    &   \multicolumn{2}{c||}{\bf En-De}    &   \multicolumn{2}{c}{\bf En-Fr}\\
  \cline{2-5}
        &   \em Step    &  \em BLEU    &   \em Step    &   \em BLEU \\
  \hline
  \hline
  Standard          &   100K    &   27.56   &   150K &   40.75\\
  ~~+ Continue      &   +20K    &   27.53   &   +20K &   40.63\\
  \hline
  ~~+ Rewind        &   +20K    &   27.70   &   +20K &   41.03\\
  \end{tabular}
  \caption{Translation performance of rewinding unimportant components. ``Step'' denotes training steps.}
  \vspace{-10pt}
  \label{tab:rewind_bleu}
\end{wraptable}

\paragraph{Rewind unimportant components and fine-tune the model.}
Recent studies have shown that rejuvenating dead neurons can improve model performance by enhancing the resource utilization of neural networks to further realize their potentials~\cite{Qiao:2019:CVPR}. Inspired by this finding, we rewound the unimportant component to the initialization values~\newcite{Frankle:2019:ICLR,Renda2020ICLR} and fine-tune them together with the other trained components for a few more steps. 
For a fair comparison, we also fine-tuned the trained Transformer model for the same number of steps.
As listed in Table~\ref{tab:rewind_bleu}, directly fine-tuning the Transformer model (``Continue'') does not outperform the standard Transformer, while the rewind technique can further improve translation performance.

\section{Related Work}

Our work is inspired by two lines of research: Interpreting transformer and  network pruning.

\paragraph{Interpreting Transformer}
Tranformer~\cite{Vaswani:2017:NeurIPS} has advanced the state of the art in various NLP tasks. Recently, there has been an increasing amount of work on interpreting specific components of Transformer, such as encoder representations~\cite{raganato2018analysis,tang:2019:emnlp,yang2019assessing}, multi-head self-attention~\cite{Li:2018:EMNLP,Voita:2019:ACL,Michel:2019:NeurIPS,Geng:2020:ACL}, and encoder attention~\cite{Jain2019AttentionIN,Li:2019:ACL,Tang:2019:RANLP}.

Closely related to our work, \newcite{Domhan:2018:ACL} investigated how much each component of Transformer matters. They revealed that self-attention is more important for the encoder side than the decoder side, and encoder attention and residual feed-forward components are key.
The key difference between their work and ours is that they evaluated the impact of individual component by retraining a model with other components, while we investigate their contribution on a trained model. In addition, we conduct more subtle analyses on components at different layers, and show other interesting findings, e.g. the lowest and uppermost layers are generally more important than intermediate layers.

\paragraph{Network Pruning}

The state-of-the-art deep neural networks are usually {\em over-parameterized}: they have much more parameters than the number of training samples.~\cite{Denton:2014:NeurIPS}. Recent study has shown that more than 90\% of the parameters can be pruned without harming the performance of neural networks~\cite{Frankle:2019:ICLR}. In response to this problem, several researchers propose pruning to extract sub-networks from the over-parameterized network with no decrease of model performance. 

Based on the granularity level of pruning, network pruning methods can be divided into weight pruning and structured pruning.
Weight pruning approaches prune the sparse weights distributed in different components~\cite{Han:2015:NeurIPS,Han:2016:ICLR}, while structured pruning removes coherent groups of weights to preserve the original structure of the network~\cite{Lin:2017:NeurIPS,HuangGao:2018:CVPR}.

In the NLP community, recent studies have shown that Transformer is over-parameterized. For example,~\newcite{Voita:2019:ACL} and~\newcite{Michel:2019:NeurIPS} showed that most self-attention heads can be dropped. \newcite{Fan:2020:LCLR} reduced Transformer depth on demand with structured dropout. Along this direction, we analyze the redundancy of Transformer on components level and reveal several interesting findings.

\section{Conclusion}

In this work we investigate the impact of individual components in Transformer on model performance. Experimental results in a couple of settings show that different components are not equally important. The decoder self-attention layers are least important, and the decoder feed-forward layers are most important. The components that are closer to the model input and output are consistently more important than the others. Upper encoder-attention layers in decoder are more important than lower encoder-attention layers. Further in-depth analyses reveal that the dropout and training data size can affect the number of unimportant components. We also find that unimportant components can be identified at early stage of training and their existence is not because of deficient training. Finally, we show that rewinding the unimportant components and then fine-tuning the Transformer model for a few more steps can further improve translation performance.

Future directions include designing better approaches to evaluate the impact of components (e.g., from the perspective of information flow), and validating our findings on other NMT architectures such as RNMT~\cite{Chen:2018:ACL} and ConvS2S~\cite{Gehring:2017:ICML}.

\bibliography{ref}
\bibliographystyle{coling}

\end{document}